\documentclass[conference]{IEEEtran}
\IEEEoverridecommandlockouts

\usepackage{cite}
\usepackage{amsmath,amssymb,amsfonts}
\usepackage{algorithmic}
\usepackage{graphicx}
\usepackage{textcomp}
\usepackage{xcolor}
\usepackage{booktabs}
\usepackage{multirow}
\usepackage{array}
\usepackage{listings}
\usepackage{url}

\lstdefinestyle{pythoncode}{
    language=Python,
    basicstyle=\footnotesize\ttfamily,
    keywordstyle=\color{blue},
    commentstyle=\color{green},
    stringstyle=\color{red},
    numbers=left,
    numberstyle=\tiny,
    stepnumber=1,
    showspaces=false,
    showstringspaces=false,
    showtabs=false,
    frame=single,
    rulecolor=\color{black},
    tabsize=2,
    captionpos=b,
    breaklines=true,
    breakatwhitespace=false
}

\begin{document}

\title{Modern Deep Learning Approaches for Cricket Shot Classification: A Comprehensive Baseline Study}


\author{\IEEEauthorblockN{Sungwoo Kang}
\IEEEauthorblockA{Department of Electrical and Computer Engineering, Korea University, Seoul 02841, Republic of Korea\\
Email: krml919@korea.ac.kr}
}

\maketitle

\begin{abstract}
Cricket shot classification from video sequences remains a challenging problem in sports video analysis, requiring effective modeling of both spatial and temporal features. This paper presents the first comprehensive baseline study comparing seven different deep learning approaches across four distinct research paradigms for cricket shot classification. We implement and systematically evaluate traditional CNN-LSTM architectures, attention-based models, vision transformers, transfer learning approaches, and modern EfficientNet-GRU combinations on a unified benchmark. A critical finding of our study is the significant performance gap between claims in academic literature and practical implementation results. While previous papers reported accuracies of 96\% (Balaji LRCN), 99.2\% (IJERCSE), and 93\% (Sensors), our standardized re-implementations achieve 46.0\%, 55.6\%, and 57.7\% respectively. Our modern SOTA approach, combining EfficientNet-B0 with a GRU-based temporal model, achieves 92.25\% accuracy, demonstrating that substantial improvements are possible with modern architectures and systematic optimization. All implementations follow modern MLOps practices with PyTorch Lightning, providing a reproducible research platform that exposes the critical importance of standardized evaluation protocols in sports video analysis research.
\end{abstract}

\begin{IEEEkeywords}
Cricket shot classification, video analysis, deep learning, transformer, sports AI, computer vision
\end{IEEEkeywords}

\section{Introduction}

Cricket, being one of the world's most popular sports with over 2.5 billion fans globally, generates enormous amounts of video content requiring automated analysis for broadcasting, coaching, and fan engagement applications. Cricket shot classification—the task of automatically identifying different batting techniques from video sequences—represents a fundamental challenge in sports video understanding that combines computer vision, temporal modeling, and domain-specific knowledge.

\subsection{Motivation}

The complexity of cricket shot classification stems from several factors: (1) high intra-class variation due to different player styles and execution quality, (2) subtle inter-class differences between similar shots like pull and hook, (3) temporal dependencies requiring understanding of shot development over time, and (4) variable video quality and camera angles in real-world scenarios. Despite these challenges, automated cricket shot classification has significant practical applications in sports broadcasting for highlight generation, coaching systems for technique analysis, and gaming applications for realistic shot synthesis.

\subsection{Research Gap}

While sports video analysis has seen substantial progress with deep learning, cricket shot classification remains under-explored compared to other sports like soccer or basketball. Existing approaches often focus on single architectural paradigms without systematic comparison, use limited datasets, or lack production-ready implementations. Furthermore, the rapid evolution of deep learning architectures—from CNN-LSTM combinations to modern transformer-based approaches—has created a need for comprehensive baseline studies that fairly compare different methodological paradigms.

\subsection{Contributions}

This paper makes the following key contributions:

\begin{enumerate}
\item \textbf{Comprehensive Baseline Study with Open-Source Implementation}: We implement, systematically evaluate, and release seven different approaches across four research paradigms, providing both rigorous experimental comparison and a complete framework for future cricket video analysis research.

\item \textbf{Performance Gap Analysis}: We reveal and quantify significant discrepancies between academic claims and practical performance, with re-implemented baselines achieving 10.6-57.7\% compared to reported 93-99.2\%, highlighting critical reproducibility challenges in sports video analysis.

\item \textbf{Modern Architecture Integration}: We introduce a systematically optimized EfficientNet-GRU architecture achieving 92.25\% accuracy, demonstrating substantial improvements possible with modern techniques and standardized evaluation protocols.
\end{enumerate}

\subsection{Paper Organization}

The remainder of this paper is organized as follows: Section II reviews related work in sports video analysis and cricket classification. Section III describes our methodology covering all seven implemented approaches. Section IV details our experimental setup including dataset, evaluation metrics, and implementation specifics. Section V presents comprehensive results and analysis. Section VI discusses findings, limitations, and implications. Section VII concludes with future research directions.

\textbf{Code Availability}: The complete source code for all baseline implementations and our proposed method is publicly available at: \url{https://github.com/hpicsk/CricShot10_Baselines}. 

\section{Related Work}
The task of cricket shot classification has been approached with various deep learning techniques, evolving from early CNN-RNN hybrids to more recent Transformer-based models. This section reviews the key literature that informs the baselines and methodologies evaluated in this study.

\subsection{Foundational Work: The CricShot10 Dataset and CNN-GRU Models}
A pivotal contribution to the field was made by Sen et al. \cite{sen2021cricshot} in their paper, "CricShotClassify: An Approach to Classifying Batting Shots from Cricket Videos Using a Convolutional Neural Network and Gated Recurrent Unit." This work is foundational as it introduced the \textbf{CricShot10 dataset}, a novel collection of 10 distinct batting shots under varied conditions, which has become a benchmark for subsequent research, including our own.

\textbf{Methodology}: The authors proposed a hybrid CNN-Gated Recurrent Unit (GRU) architecture. Their research followed a progressive enhancement strategy, starting with a custom CNN-GRU model and then exploring a Dilated CNN-GRU to increase the receptive field. The most successful approach involved transfer learning with a pretrained VGG16 network as the feature extractor. They systematically experimented with fine-tuning different parts of the VGG16 network, investigating models where all layers were frozen, only the final 4 layers were trainable, and the final 8 layers were trainable.

\textbf{Dataset and Results}: On their newly created CricShot10 dataset, using 15 frames per video, they reported a peak accuracy of \textbf{93\%} with the VGG16-GRU model where the final 8 layers were fine-tuned. This work established a strong baseline for using transfer learning in cricket shot classification.

\subsection{Early Approaches with Modified LRCNs}
Among the pioneering efforts, Kumar et al. (published as Balaji et al. in some versions) \cite{kumar2019video} adapted the Long-term Recurrent Convolutional Network (LRCN) architecture for this task. Their paper, "Video based cricket shot classification using modified LRCN approach," combined a CNN for spatial feature extraction with an LSTM for temporal modeling.

\textbf{Methodology}: A key characteristic of their method was an aggressive preprocessing pipeline designed for computational efficiency. Video frames were cropped to a fixed region of interest `[120:600, 360:920]` and significantly down-sampled to a 64x64 resolution. The model itself consisted of a relatively shallow 4-layer CNN followed by a single LSTM layer.

\textbf{Dataset and Results}: The authors reported a high accuracy of \textbf{96\%}. However, this was achieved on a reduced dataset comprising only 5 shot classes, rather than the full 10 classes available in datasets like CricShot10.

\subsection{Comparative Analysis of Modern Deep Learning Paradigms}
More recently, Bhat et al. \cite{bhat2024classification} conducted a comparative study titled, "Classification of Cricket Shots from Cricket Videos using Deep Learning Models." This work is notable for its direct comparison of three different deep learning philosophies on a consistent dataset.

\textbf{Architectures Compared}:
\begin{enumerate}
    \item \textbf{CNN + RNN}: A conventional architecture using a 4-layer CNN followed by a 2-layer GRU for temporal analysis.
    \item \textbf{Attention Networks}: An enhanced model that employed a Bidirectional LSTM with an attention mechanism to focus on the most relevant frames in a sequence.
    \item \textbf{Vision Transformer (ViT) + RNN}: A hybrid approach that used a Vision Transformer to extract spatial features from each frame, which were then passed to a GRU to model temporal relationships.
\end{enumerate}

\textbf{Dataset and Results}: The study was performed on frames resized to 100x100 pixels with a sequence length of 25. The authors reported exceptionally high accuracies across the board: \textbf{99.2\%} for the CNN+RNN model, \textbf{99.19\%} for the Attention Network, and \textbf{98.9\%} for the ViT-based model. These results represent some of the highest claims in the literature, though our study investigates their reproducibility under a standardized framework.

\section{Methodology}

To establish a comprehensive and reproducible benchmark for cricket shot classification, we re-implemented and systematically evaluated seven distinct deep learning models from prior work. These models are grouped into four paradigms, representing different architectural philosophies and historical approaches found in the literature. This section details the preprocessing pipelines and the specific architecture of each model.

\subsection{Preprocessing Pipeline Comparison}
The choice of preprocessing pipeline has a profound impact on model performance, affecting both the information available to the model and its computational efficiency. The four paradigms studied employ vastly different strategies, as summarized in Table \ref{tab:preprocessing}.

\begin{table}[h]
\centering
\caption{Preprocessing Pipeline Comparison}
\label{tab:preprocessing}
\resizebox{\columnwidth}{!}{
\begin{tabular}{lcccc}
\toprule
\textbf{Specification} & \textbf{Kumar et al. [19]} & \textbf{Bhat et al. [24]} & \textbf{Sen et al. [21]} & \textbf{Proposed Method} \\
\midrule
Input Resolution & $64\times64$ & $100\times100$ & $180/224\times224$ & $224\times224$ \\
Sequence Length & 25 frames & 25 frames & 15 frames & 30 frames \\
Frame Sampling & Evenly Spaced & Evenly Spaced & Random & Uniform \\
Cropping & Fixed Crop & Full Frame & Full Frame & Aspect-Preserving \\
Normalization & [0,1]+ImageNet & [0,1] Range & [0,1] Range & ImageNet Std \\
\bottomrule
\end{tabular}%
}
\end{table}

\begin{itemize}
    \item \textbf{Kumar et al.}: This paradigm prioritizes computational efficiency via an aggressive preprocessing strategy. It uses a hard-coded crop to a specific region of interest and a significant downsampling to a $64\times64$ resolution. While fast, this approach risks discarding crucial spatial information and is not robust to variations in camera positioning.
    \item \textbf{Bhat et al.}: This approach represents a more conventional academic methodology. The full frame is resized to a $100\times100$ resolution, preserving the entire scene context but at a lower fidelity than modern approaches.
    \item \textbf{Sen et al.}: This work uses a shorter sequence length of 15 frames and introduces random frame sampling as a data augmentation technique. This can improve model generalization by introducing variability during training, but it may also fail to capture the full temporal evolution of a shot. The resolution is higher, at $180\times224$ or $224\times224$, depending on the specific model.
    \item \textbf{Proposed Method}: Our approach is aligned with modern best practices for video analysis. We use a $224\times224$ resolution, which is standard for many pretrained ImageNet models like EfficientNet. A longer sequence of 30 frames is sampled uniformly to provide richer temporal context. Crucially, we employ aspect-ratio-preserving resizing followed by padding to prevent image distortion, ensuring that the geometric integrity of the batting form is maintained.
\end{itemize}

\subsection{Model Architectures}

\subsubsection{Paradigm 1: Adaptation Study (Kumar et al.)}
This approach adapts the Long-term Recurrent Convolutional Network (LRCN) architecture, originally proposed by Donahue et al. \cite{donahue2015long}, for the task of cricket shot classification. The model is designed for high computational efficiency.
\begin{itemize}
    \item \textbf{Spatial Feature Extractor}: A lightweight 4-layer Convolutional Neural Network (CNN) processes each frame independently. The architecture employs aggressive $4\times4$ max-pooling layers to quickly reduce spatial dimensions and a high dropout rate of 0.4 for regularization.
    \item \textbf{Temporal Modeler}: The sequence of frame-level features extracted by the CNN is then fed into a single-layer Long Short-Term Memory (LSTM) network with 32 hidden units. The LSTM is responsible for modeling the temporal dynamics of the shot.
    \item \textbf{Classifier}: A final fully-connected layer maps the LSTM output to the 10 shot classes.
\end{itemize}
The overall architecture is intentionally shallow to facilitate rapid training and inference, but this comes at the cost of reduced feature representation capacity.

\subsubsection{Paradigm 2: Comparative Analysis (Bhat et al.)}
This body of work provides a direct comparison of three distinct deep learning philosophies, all operating on $100\times100$ frames.
\begin{itemize}
    \item \textbf{CNN+RNN}: This model consists of a 4-layer CNN for spatial feature extraction, which produces a flattened feature vector of 9,216 dimensions per frame. These features are then processed by a 2-layer Gated Recurrent Unit (GRU) network with 512 hidden units to capture temporal patterns.
    \item \textbf{Attention Network}: To improve upon the standard RNN approach, this model incorporates an attention mechanism. It uses a Bidirectional LSTM (Bi-LSTM) to generate frame-level features that encode context from both past and future frames. An attention layer then computes a weighted sum of these features, allowing the model to dynamically focus on the most salient frames in the sequence before making a classification.
    \item \textbf{Vision Transformer (Hybrid)}: This hybrid model combines a Vision Transformer (ViT) with an RNN. A ViT with a 6-layer Transformer encoder processes each frame's 20x20 patches for spatial feature extraction. The resulting sequence of frame embeddings is then passed to a GRU to model their temporal relationships.
\end{itemize}

\subsubsection{Paradigm 3: Progressive Enhancement (Sen et al.)}
This suite of models demonstrates a systematic, progressive enhancement strategy, starting with a custom architecture and moving towards transfer learning.
\begin{itemize}
    \item \textbf{Custom CNN-GRU}: The baseline is a custom-designed 5-layer CNN paired with a GRU, representing a standard deep learning approach for video classification.
    \item \textbf{Dilated CNN-GRU}: To enhance the CNN's ability to capture spatial context without increasing the number of parameters, this model replaces standard convolutions with dilated convolutions. This modification expands the receptive field of the convolutional filters, allowing them to capture information from a larger area of the input frame.
    \item \textbf{VGG16-GRU}: The most successful model from this paradigm utilizes transfer learning. A VGG16 network, pretrained on the ImageNet dataset, serves as a powerful feature extractor. The authors experimented with different fine-tuning strategies: (1) freezing the entire VGG16 backbone, (2) fine-tuning only the final 4 layers, and (3) fine-tuning the final 8 layers. This approach leverages the rich visual features learned from a large-scale dataset.
\end{itemize}

\subsubsection{Paradigm 4: Modern Optimization (Proposed Method)}
Our proposed method is a novel architecture designed to maximize performance by integrating state-of-the-art components and systematic hyperparameter optimization.
\begin{itemize}
    \item \textbf{Spatial Feature Extractor}: We employ a pretrained EfficientNet-B0 as the CNN backbone. EfficientNet models are known for their superior accuracy and efficiency, achieved through a compound scaling method that uniformly scales network depth, width, and resolution. It extracts a 1,280-dimensional feature vector for each frame.
    \item \textbf{Temporal Modeler}: We employ a 2-layer bidirectional Gated Recurrent Unit (GRU) to model temporal dynamics. The output of the GRU is then passed to a temporal attention mechanism, which learns to assign importance to different frames before making a final prediction.
    \item \textbf{Aggregation and Classification}: The temporal attention mechanism produces a single context vector representing the entire video. This vector is then passed to a final linear layer for classification.
    \item \textbf{Systematic Optimization}: To find the optimal configuration, we utilized the Optuna framework for hyperparameter optimization. This systematic search tuned key parameters such as learning rate, weight decay, and architectural choices like the number of GRU layers and hidden dimensions, ensuring that the final model operates at its peak potential.
\end{itemize}
This combination of an efficient backbone, a powerful recurrent temporal model, and rigorous optimization defines our state-of-the-art approach.
\section{Experimental Setup}

\subsection{Dataset Description}

\textbf{CricShot10 Dataset}: All models were trained and evaluated on the CricShot10 dataset, which was sourced from the original authors and is a comprehensive collection containing 10 distinct batting techniques: Cover Drive, Defensive Shot, Flick Shot, Hook Shot, Late Cut, Lofted Shot, Pull Shot, Square Cut, Straight Drive, and Sweep Shot

\textbf{Dataset Statistics}:
\begin{itemize}
\item Total videos: 1,894 clips
\item Average duration: 3.2 seconds per clip
\item Resolution: $1280\times720$ pixels
\item Class distribution: Approximately balanced (180-200 samples per class)
\end{itemize}

\textbf{Data Splits}: We employ a stratified split using a fixed random seed (27) to ensure consistent class distribution across sets and reproducibility. The dataset is divided into 70\% for training (1,320 samples), 15\% for validation (284 samples), and 15\% for testing (284 samples). This stratified approach maintains balanced representation of all 10 shot classes across each split.

\begin{table}[h]
\centering
\caption{Detailed Dataset Split Distribution by Class}
\label{tab:dataset_split}
\begin{tabular}{lccc}
\toprule
\textbf{Shot Class} & \textbf{Training Set} & \textbf{Validation Set} & \textbf{Test Set} \\ 
\midrule
Cover & 131 & 29 & 28 \\ 
Defense & 134 & 29 & 29 \\ 
Flick & 127 & 27 & 27 \\ 
Hook & 127 & 27 & 27 \\ 
Late Cut & 127 & 27 & 28 \\ 
Lofted & 138 & 30 & 30 \\ 
Pull & 125 & 27 & 27 \\ 
Square Cut & 140 & 30 & 30 \\ 
Straight & 135 & 29 & 29 \\ 
Sweep & 136 & 29 & 29 \\ 
\midrule
\textbf{Total} & \textbf{1,320} & \textbf{284} & \textbf{284} \\ 
\bottomrule
\end{tabular}
\end{table}

\subsection{Evaluation Metrics}

Accuracy, and weighted Precision, Recall, and F1-Score are used to evaluate model performance.

\subsection{Implementation Details}
All models were implemented strictly following the details described in the respective papers. Training and evaluation were performed on a single NVIDIA A100 GPU (40GB memory).

\section{Results and Analysis}

\subsection{Overall Performance Comparison}
The results of our comprehensive evaluation reveal a clear performance hierarchy and expose a significant gap between academic claims and practical performance on a unified benchmark.

\begin{table}[h]
\centering
\caption{Overall Performance Comparison on CricShot10}
\label{tab:results}
\resizebox{\columnwidth}{!}{
\begin{tabular}{lccccc}
\toprule
\textbf{Model Architecture} & \textbf{Accuracy} & \textbf{Precision} & \textbf{Recall} & \textbf{F1-Score} & \textbf{Reference} \\ 
\midrule
\textbf{EfficientNet-B0 + GRU} & \textbf{92.25\%} & \textbf{92.27\%} & \textbf{92.25\%} & \textbf{92.13\%} & \textbf{This work} \\ 
Dilated CNN-GRU [Sen et al.] & 57.67\% & 58.20\% & 57.67\% & 57.34\% & \cite{sen2021cricshot} \\ 
Custom CNN-GRU [Sen et al.] & 55.82\% & 56.30\% & 55.82\% & 54.05\% & \cite{sen2021cricshot} \\ 
CNN-RNN [Bhat et al.] & 55.63\% & 58.24\% & 55.63\% & 55.56\% & \cite{bhat2024classification} \\ 
VGG16-GRU Final8 [Sen et al.] & 55.29\% & 58.38\% & 55.29\% & 56.00\% & \cite{sen2021cricshot} \\ 
VGG16-GRU Frozen [Sen et al.] & 48.94\% & 48.73\% & 48.94\% & 48.20\% & \cite{sen2021cricshot} \\ 
LRCN [Kumar et al.] & 46.03\% & 47.26\% & 46.03\% & 45.90\% & \cite{kumar2019video} \\ 
Attention Network [Bhat et al.] & 40.49\% & 39.10\% & 40.49\% & 37.78\% & \cite{bhat2024classification} \\ 
VGG16-GRU Final4 [Sen et al.] & 26.46\% & 18.63\% & 26.46\% & 16.93\% & \cite{sen2021cricshot} \\ 
ViT-GRU Hybrid [Bhat et al.] & 10.56\% & 1.12\% & 10.56\% & 2.02\% & \cite{bhat2024classification} \\ 
\bottomrule
\end{tabular}%
}
\end{table}

\textbf{Key Findings}:
\begin{enumerate}
    \item \textbf{SOTA Model Dominance}: Our proposed method significantly outperforms all other models, achieving 92.25\% accuracy. This validates the effectiveness of combining a modern backbone with a GRU-based temporal model, demonstrating a substantial improvement over traditional approaches.
    \item \textbf{The Performance Gap}: There is a stark contrast between the accuracies reported in the original Bhat et al. (99.2\%) and Sen et al. (93\%) papers and the results from our re-implementation (55.6\% and 57.7\%, respectively). This highlights the critical importance of standardized, open-source benchmarking.
    \item \textbf{Architectural Insights}: The Sensors dilated CNN-GRU approach proves most effective among baseline methods (57.7\%), while attention-based and vision transformer approaches show surprisingly poor performance (40.5\% and 10.6\% respectively) under standardized evaluation conditions.
\end{enumerate}

\section{Discussion}

\subsection{Architectural Insights and the Performance Gap}
Our results strongly suggest that modern architectures, when properly optimized, are superior for complex video tasks. The SOTA model's success is attributable to the powerful spatial features from EfficientNet and the GRU's ability to effectively model temporal dependencies, further enhanced by a temporal attention mechanism.

The performance gap between our results and those reported in prior work is a critical finding. This discrepancy is likely due to differences in dataset splits, evaluation code, or minor implementation details not specified in the papers. It underscores the necessity of open-sourcing code and using standardized benchmarks to ensure the field progresses on solid footing. Our work provides this necessary baseline.

\section{Conclusion and Future Work}

This paper presented the first comprehensive, unified baseline study for cricket shot classification, re-implementing and evaluating seven models across four distinct research paradigms. Our work makes three primary contributions. First, we introduce a novel, state-of-the-art model combining EfficientNet and a GRU-based temporal model, which achieves a new benchmark accuracy of 92.25\% on the CricShot10 dataset. Second, we highlight and quantify the significant gap between academic-reported accuracies and practical performance, emphasizing the need for standardized evaluation. Third, we provide a fully open-source framework with production-ready implementations that will serve as a robust platform for future research.

\end{document}